\documentclass[lettersize,journal]{IEEEtran}
\usepackage{amsmath,amsfonts,amssymb}
\usepackage{enumerate}
\usepackage{algorithmic}
\usepackage{algorithm}
\usepackage{array}
\usepackage[caption=false,font=normalsize,labelfont=sf,textfont=sf]{subfig}
\usepackage{textcomp}
\usepackage{stfloats}
\usepackage{url}
\usepackage{verbatim}
\usepackage{graphicx}
\usepackage{cite}
\usepackage{color}
\usepackage{multirow}
\usepackage{booktabs}
\usepackage{lineno,hyperref}
\begin{document}

\title{Multimodal Informative ViT: Information Aggregation and Distribution for Hyperspectral and LiDAR Classification}


\author{Jiaqing Zhang,
	Jie Lei,~\IEEEmembership{Member,~IEEE},
	Weiying Xie,~\IEEEmembership{Senior Member,~IEEE},
   Geng Yang, 
   Daixun Li,
	Yunsong Li,~\IEEEmembership{Member,~IEEE}

    

\thanks{This work was supported in part by the National Natural Science Foundation of China under Grant 62071360. (Corresponding~authors: Jie Lei)
Jiaqing Zhang, Jie Lei, Weiying Xie, Daixun Li, Geng Yang, and Yunsong Li are with the State Key
Laboratory of Integrated Services Networks, Xidian University, Xi'an 710071,
China (e-mail: jqzhang\underline{ }2@stu.xidian.edu.cn; jielei@mail.xidian.edu.cn; wyxie@xidian.edu.cn; gengyang@stu.xidian.edu.cn; ldx@stu.xidian.edu.cn;ysli@mail.xidian.edu.cn).
}
}

\markboth{Journal of \LaTeX\ Class Files,~Vol.~14, No.~8, November~2021}%
{Zhang \MakeLowercase{\textit{et al.}}: Multimodal Informative ViT: Information Aggregation and Distribution for Hyperspectral and LiDAR Classification}


\maketitle

\begin{abstract}
In multimodal land cover classification (MLCC), a common challenge is the redundancy in data distribution, where irrelevant information from multiple modalities can hinder the effective integration of their unique features. To tackle this, we introduce the Multimodal Informative Vit (MIVit), a system with an innovative information aggregate-distributing mechanism. This approach redefines redundancy levels and integrates performance-aware elements into the fused representation, facilitating the learning of semantics in both forward and backward directions. MIVit stands out by significantly reducing redundancy in the empirical distribution of each modality's separate and fused features. It employs oriented attention fusion (OAF) for extracting shallow local features across modalities in horizontal and vertical dimensions, and a Transformer feature extractor for extracting deep global features through long-range attention. We also propose an information aggregation constraint (IAC) based on mutual information, designed to remove redundant information and preserve complementary information within embedded features. Additionally, the information distribution flow (IDF) in MIVit enhances performance-awareness by distributing global classification information across different modalities' feature maps. This architecture also addresses missing modality challenges with lightweight independent modality classifiers, reducing the computational load typically associated with Transformers. Our results show that MIVit's bidirectional aggregate-distributing mechanism between modalities is highly effective, achieving an average overall accuracy of 95.56\% across three multimodal datasets. This performance surpasses current state-of-the-art methods in MLCC. The code for MIVit is accessible at https://github.com/icey-zhang/MIViT.
\end{abstract}


\section{Introduction}
\label{sec:intro}
\IEEEPARstart{L}{and} Cover Classification (LCC) is a key area in remote sensing image processing, focusing on identifying and categorizing land types or cover classes on Earth's surface areas \cite{xie2020multiscale,liu2021global,wang2022am3net}. Advances in satellite technology have enabled the capture of multi-source imagery from various satellite sensors, leading to the growth of multimodal fusion in remote sensing image analysis \cite{bandara2022hypertransformer,xu2021deep,zhao2023cddfuse}. Each data source, such as Hyperspectral Imaging (HSI) and LiDAR, provides unique insights into the same geographical area \cite{mohla2020fusatnet,10036472}, enriching LCC by offering more comprehensive representations of scenes and objects in remote sensing images \cite{wang2023mutually}. Despite these advancements, effectively utilizing information from various multimodal sources to improve the accuracy and reliability of LCC remains a significant challenge.

\begin{figure}[tp]
    \centering
    \includegraphics[width=\linewidth]{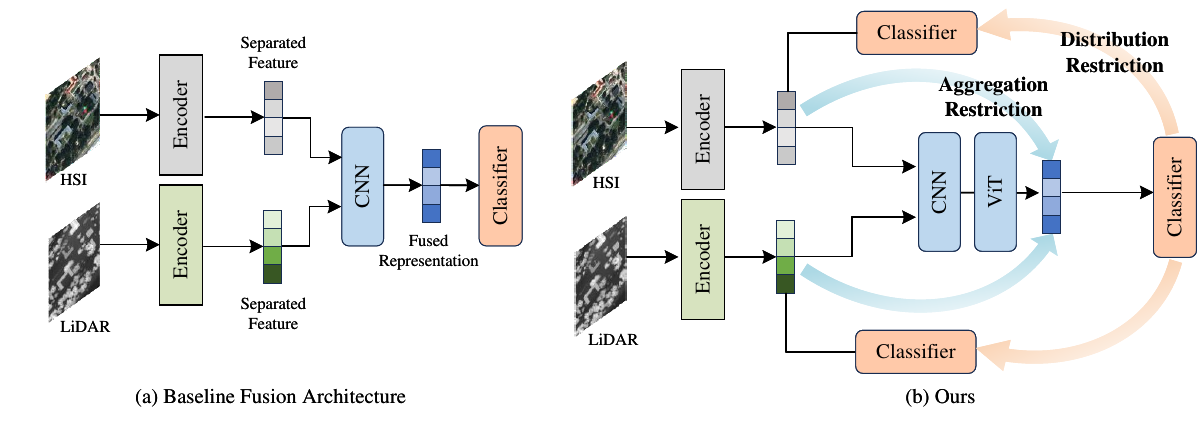}
    \caption{Existing MLCC method \textit{vs}. ours MIViT.}
    \label{fig:question}
    \vspace{-0.1in}
\end{figure}
In recent years, many methods have been developed to address the challenges of multimodal land cover classification (MLCC) \cite{roy2023multimodal,ding2022global, lu2023coupled}. A common pipeline that demonstrated promising results utilizes CNN-based feature extraction in an end-to-end manner \cite{chen2017deep,xu2017multisource,hong2020deep, zhao2020joint,wu2021convolutional,10105921}. For instance, Chen \textit{et al.} \cite{chen2017deep} initially employed deep neural networks to extract features from HSI and LiDAR for precise classification. Xu \textit{et al.} designed a dual-branch CNN framework, where one branch was dedicated to extracting spectral-spatial features from HSI, and another was designed for LiDAR data feature extraction. Similarly, Dong \textit{et al.} \cite{dong2022multibranch} developed the MB2FscgaNet, a multi-branch fusion network emphasizing self- and cross-guided attention to refine HSI and LiDAR classification, thereby enhancing the accuracy of spectral, spatial, and elevation feature estimations. Additionally, Wu \textit{et al.} \cite{wu2021convolutional}introduced the CCR-Net, incorporating a novel cross-channel reconstruction module, and Zhang \textit{et al.} \cite{zhang2021spectral} devised a spectral-spatial fractal residual convolutional neural network with data balance augmentation to address dataset-specific challenges such as imbalanced categories and limited labeled data. Li \textit{et al.} \cite{li2023fedfusion} first considers on-orbit fusion with limited computing resources and provides a strategy manifold structure for multi-modal fusion transmission.

These methodologies generally involve two distinct phases: the fusion of multimodal data through feature extractors to create latent mapping representations, followed by the classification of these representations, as illustrated in Fig. \ref{fig:question}a. Despite their effectiveness, these methods exhibit certain limitations. Firstly, conventional CNNs, being context-independent, primarily extract local information within a limited receptive field and struggle with global information acquisition \cite{liang2021swinir}. Secondly, the opaque internal mechanisms of CNNs often lead to the insufficient and redundant extraction of cross-modality features. Thirdly, in single-modal feature extraction, the distance from the initial feature extraction to the final decision classifier often results in a significant loss of task-relevant knowledge \cite{zhang2019your}. 
To address these shortcomings, Transformer-based algorithms \cite{yao2023extended,roy2023multimodal,zhao2022fractional,zhao2022joint} have been introduced, offering advantages in capturing global long-range sequence information. However, they also face challenges, particularly in terms of the high computational resources required \cite{zhang2023essaformer,2023arXiv231109520L}. In light of these issues, our paper focuses on reducing information redundancy to enhance feature fusion and foster performance-aware learning and classification discrimination.

Aiming to improve the controllability and interpretability of multimodal fusion, we propose the Multimodal Informative ViT (MIViT) algorithm, as shown in Fig. \ref{fig:question}b. This approach initially transforms multimodal data into low-dimensional separated features using CNNs for local context and efficiency, and Transformers for global attention and long-range dependency, particularly in addressing the LCC task. We utilize an Oriented Attention Fusion (OAF) based on CNN to extract cross-modality shallow local context features in both horizontal and vertical dimensions. Furthermore, the Transformer feature extractor ViT \cite{dosoViTskiy2020image} is designed to leverage global attention and long-range dependency for handling deep global features, enhancing the overall effectiveness and accuracy of the LCC process.

Inspired by information theory, MIViT integrates an iterative refinement mechanism for information aggregation and distribution. It comprises two branches: the forward aggregation branch and the backward distribution branch. In the forward aggregation branch, mutual information (MI) \cite{bramon2011multimodal} is utilized to mathematically quantify and define the redundancy present in feature information. Building upon this concept, we have developed an Information Aggregation Constraint (IAC), which is grounded in MI. The primary aim of IAC is to selectively filter out redundant and irrelevant information from tasks, thereby facilitating the formation of compact, yet highly reliable representations.

The backward distribution branch plays a crucial role in enhancing the granularity of classification results. It takes the globally fused features and effectively distributes them across separated feature maps. This is achieved through the Information Distribution Flow (IDF), which is based on the principles of self-distillation (SD) \cite{zhang2019your}. The IDF is instrumental in reducing discrepancies between different classification maps, thereby significantly improving the perceptual capabilities of individual modalities. This unique approach allows for a more holistic and comprehensive construction of multimodal fusion representations. Notably, this distillation method permits the output of classification results through shallow classifiers for each modality, independent of the presence of multiple modalities. This feature is especially significant as it leverages a lightweight CNN architecture, enabling long-range knowledge transfer similar to that of deep Transformer networks, but without the associated computational burden.

Our contributions to this work are multifaceted and can be summarized as follows:
\begin{itemize}
\item We introduce the MIViT network, featuring an innovative information aggregate-distributing mechanism. This mechanism is adept at extracting and fusing both global and local features, thereby facilitating a dynamic and interactive flow of information throughout the multimodal fusion process. The result is the generation of compact representations that are acutely aware of performance dynamics.

\item Aiming to remove less task-related redundant information, We formulate an information aggregation constraint based on mutual information. This constraint is a pioneering step in redefining the redundancy that exists between multimodal data and fusion representations, and it is integrated as a regularization term within the loss function.

\item We design an information distribution flow, grounded in self-distillation constraints, allowing for the effective dissemination of global fusion information to feature maps of various modalities. This not only enhances the performance awareness of individual modalities but also leads to the creation of more comprehensive and enriched fusion representations.

\item Our research marks the first exploration into the realm of missing modality learning via knowledge distillation. We delve into the capabilities of lightweight CNN models in handling missing modalities, showcasing their potential to mimic the learning efficiency of Transformers while evading the extensive computational demands typically associated with Transformer architectures.

\end{itemize}
\begin{figure*}[htpb]
	\centering
	\includegraphics[width=\linewidth]{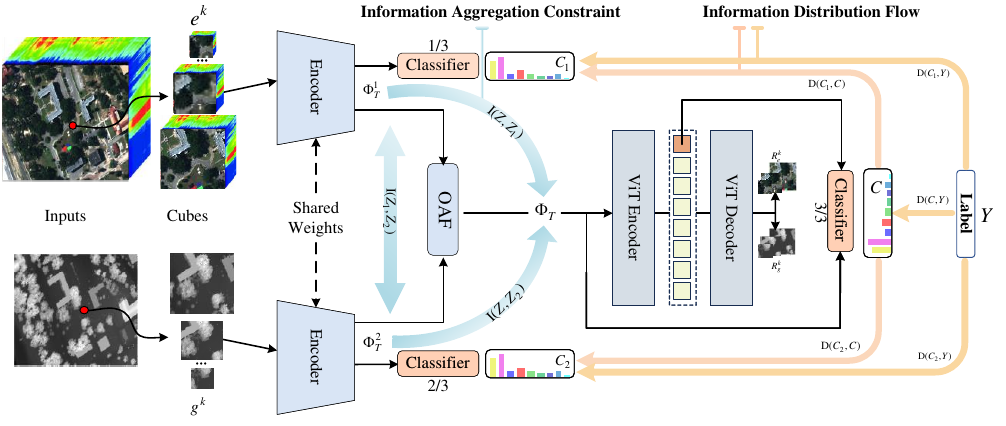}
	\caption{The detailed frame of our proposed method. The different modalities are firstly fed into an Alignment Encoder (AE) to capture the aligned separated features ($\Phi_T^1$ and $\Phi_T^2$) from shallow to deep levels. Subsequently, we fuse the separated features to effectively model the complementary information of each mode by Oriented Attention fusion (OAF). To explicitly encourage complementary learning, eliminate information redundancy, and enhance the performance perception capability of multi-classifiers, we impose both the Information Aggregation Constraint (IAC) and Information Distribution Flow (IDF) over the separated and fused representations.}
	\label{fig:framenew}
\end{figure*}

\section{Related Work}
\label{sec:epf}
This section provides an overview of significant advancements in deep learning-based fusion approaches and multimodal learning with mutual information, as employed in MIViT.
\subsection{Deep learning-based Fusion methods}
Nowadays, with the success of deep learning (DL) methods in various fields, the DL methods have also begun to be utilized in multi-modal learning and have made great progress.
The integration of deep learning (DL) methods into multimodal learning has marked a significant shift in this field, leading to substantial progress. The fundamental premise of multimodal learning is the existence of both shared and distinct information across different modalities \cite{zhao2023cddfuse}. In the context of Land Cover Classification (LCC), datasets comprising Hyperspectral Images (HSIs), LiDAR images, and SAR images exhibit shared semantic information. While HSI predominantly encodes spectral information, LiDAR and SAR are more focused on geometric details, like the relative elevation or localization of objects. This distinction between spectral and geometric information highlights the uniqueness of each modality. Popular approaches to achieve multimodal learning include early fusion, late fusion \cite{han2022trusted}, , and cross fusion \cite{wang2023nearest,zhao2022joint,hang2020classification}.Early fusion involves direct concatenation of modalities, whereas late fusion processes each modality separately, combining them only at the output layer. Cross fusion, as demonstrated in studies \cite{mohla2020fusatnet, zhao2022joint}, effectively strengthens the interaction and semantic correlation between different modalities.

Existing methods, while adept at fusing for multimodal learning, often do not explicitly address how the CNN network facilitates effective multimodal learning of fused representation. Our approach, with its focus on aggregation and distribution constraints, aims to minimize the redundancy of appearance and geometric features, thereby promoting a more performance-aware and efficient multimodal learning process.

\subsection{Multi-modal learning with mutual information}
In machine learning, mutual information is frequently used as a criterion to either encourage or limit dependencies between variables in a range of tasks \cite{Zhao2023,wu2023infoctm,su2023towards}. In the realm of multimodal learning, deep mutual information maximization methods have been proposed for applications like cross-clustering \cite{mao2021deep}. These methods focus on preserving shared information across modalities while discarding extraneous individual modal information in an end-to-end manner. Another approach maximizes mutual information between 3D objects and their geometrically transformed versions to refine representations \cite{sanghi2020info3d}. Contrarily, mutual information minimization is employed to explicitly foster multimodal information learning, as seen in the interaction between RGB images and depth data \cite{zhang2021rgb}. Similarly, mutual information has been incorporated as an effective representation regularizer in tasks like Pan-sharpening \cite{zhou2022mutual}. 

The information redundancy in the feature extraction of the various modalities naturally exists in MLCC task \cite{li2022asymmetric} which leads to the limit of the effective feature fusion. To address this, we propose the minimization of mutual information between two modalities to reduce specific information redundancy. Additionally, we aim to maximize mutual information between the fusion features and other isolated modality features. This approach not only captures sufficient complementary information but also enhances the overall performance of fusion features. Importantly, our focus extends beyond just learning representations of different modalities; we also concentrate on the fused representation itself. This aspect of our work sets it apart from previous studies, emphasizing our unique approach to multimodal learning with mutual information.

\section{Method}
\label{sec:proposed}
In this section, we give the mathematical definition of task-relevant and task-irrelevant information for multimodal learning and fusion in the LCC task. We then detail the process of achieving compact representation and describe the implementation of our information aggregate-distributing mechanism.

\subsection{Problem Definition}
The essence of LCC lies in the precise categorization of each pixel in an image into a specific land cover category. In the context of multimodal LCC, this involves the complex processes of multimodal learning and fusion. Given $n$ heterogeneous remote sensing modalities, denoted as $X_n \in \mathbb{R}^{h \times w \times b}$, the challenge is to extract a compact and reliable representation $Z$ from the input image $X$. This representation is then used to map into a new probability distribution $C \in \mathbb{R}^{h \times w \times c}$, indicating the likelihood of each pixel belonging to different classes. In this scenario, $h$, $w$, $b$, and $c$ represent height, width, number of channels, and classification categories, respectively. The success of this task hinges on the efficient coupling of information and the perception of performance within the latent fusion representation, which significantly influences the accuracy of the final classification results. In our approach, multimodal LCC involves two key processes: eliminating task-irrelevant information $r^*$ to reduce redundancy between fusion modes and enhancing task-relevant information $s^*$. This dual approach ensures that the classification results of fusion representations and individual modal features mutually constrain each other during the classification process. The problem can be defined as:
\begin{equation}
	\begin{aligned}
	\text{min} &r^* = \sum_{i=1}^n \sum_{j > i}^n \operatorname{I} \left (Z_i, Z_j\right)-\sum_{i=1}^n \operatorname{I} \left(Z, Z_i\right) \\
	\text{max} &s^* = -(\sum_{i=1}^n \operatorname{D} \left(C_i, C\right) + \sum_{i=1}^n \operatorname{D} \left(C_i, Y\right)+\operatorname{D} \left(C, Y\right))  \\
	\end{aligned}
\end{equation}
where $Z_i$ are probability distributions of compressed embedding of modalities $X_i$, so minimizing mutual information $\operatorname{I}(Z_i, Z_j)$ can eliminate the redundant information between $X_i$ and $X_j$. $Z$ represents the fused representation and the complementary information of multiple modalities by maximizing mutual information between $Z$ and $Z_i$. The similarity distribution distance \cite{ou2021sdd} between the two classification results after a softmax operation is defined as $\operatorname{D}(\cdot)$. The maximum value of a negative distance reflects a greater degree of similarity between distributions. To simplify the expression, we set $n=2$ in the subsequent paper.

\subsection{Multimodal Informative ViT}
\subsubsection{Alignment Encoder}
We employ a multi-scale shared encoder to align latent feature distribution. This encoder utilizes a convolutional block to project multi-modal images to modality-aware maps, extracting multi-scale shallow features for each modality efficiently. The encoder splits $N$ types of cubes from the original images, denoted as $\{e^k\}_{k=1}^N$ and $\{g^k\}_{k=1}^N$, for their respective modalities. Two branches, $\mathcal{P(\cdot)}$ and $\mathcal{Q(\cdot)}$, help obtain multi-scale shallow features:
\begin{equation}
  \label{ae1}
	\Phi_P^k=\mathcal{P}(e^k), \Phi_Q^k=\mathcal{Q}(g^k).
\end{equation}

The features, $\Phi_P^k$ and $\Phi_Q^k$, are then mapped to the same spatial dimension using a convolutional block and max-pooling operation. Assuming that the embedding functions are $\{\mathcal{F(\cdot)}\}_{k=1}^N$ and $\{\mathcal{G(\cdot)}\}_{k=1}^N$. The base features extracted from the shared features can be represented as:
\begin{equation}
\label{ae2}
	\Phi_F^k=\mathcal{F}^k(\Phi_P^k), \Phi_G^k=\mathcal{G}^k(\Phi_Q^k).
\end{equation}
The base features are adaptively linearly weighted to achieve multi-scale feature fusion:
\begin{equation}
\label{ae3}
	\Phi_T^1 = {\alpha}_k \Phi_F^k,	\Phi_T^2 = {\alpha}_k \Phi_G^k,
\end{equation}
where ${\alpha}_k$ representing the adaptive weighting coefficient, learnable and updatable by the network. 
\begin{figure}[htpb]
	\centering
	\includegraphics[width=\linewidth]{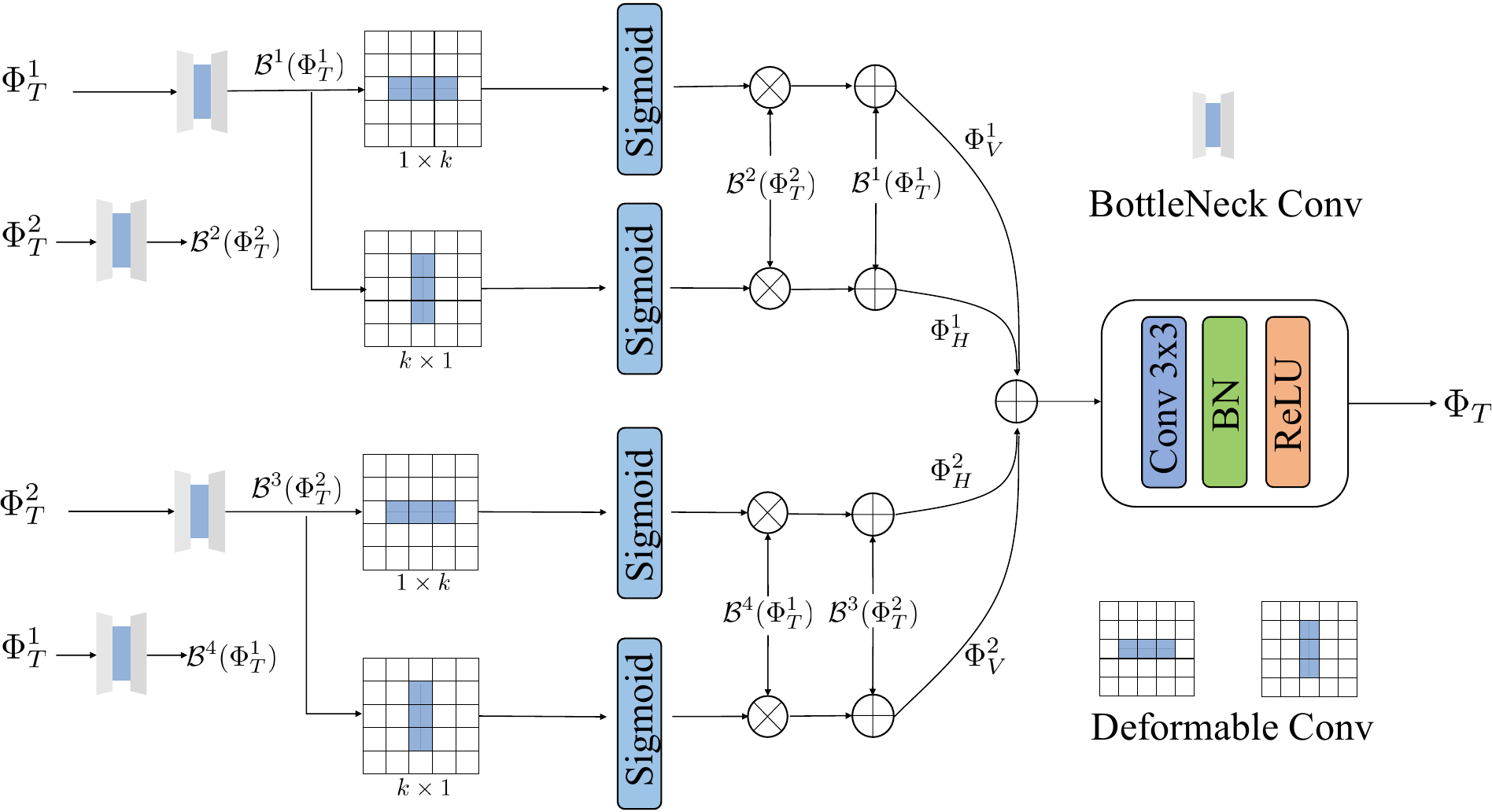}
	\caption{The structure of oriented attention fusion (OAF) module.}
	\label{fig:fusion}
 \vspace{-0.1in}
\end{figure}

\subsubsection{Oriented Attention Fusion (OAF)}
The OAF module fuses multi-modal features, filtering redundant information through two-orientation attention aggregation. Each attention module in OAF generates an attention map along one direction (horizontal or vertical) and modulates one modality's features. The process can be written as:
\begin{equation}
\label{oaf_1}
		\begin{aligned}
	\Phi_H^1 &= \mathcal{S}(\mathcal{E}_{1 \times k}^1(\mathcal{B}^1(\Phi_T^1))*\mathcal{B}^2(\Phi_T^2)+\mathcal{B}^1(\Phi_T^1),\\
	\Phi_V^1 &= \mathcal{S}(\mathcal{E}_{k \times 1}^1(\mathcal{B}^1(\Phi_T^1))*\mathcal{B}^2(\Phi_T^2)+\mathcal{B}^1(\Phi_T^1),\\
	\Phi_H^2 & = \mathcal{S}(\mathcal{E}_{1 \times k}^2(\mathcal{B}^3(\Phi_T^2))*\mathcal{B}^4(\Phi_T^1)+\mathcal{B}^3(\Phi_T^2),\\
	\Phi_V^2 & = \mathcal{S}(\mathcal{E}_{k \times 1}^2(\mathcal{B}^3(\Phi_T^2))*\mathcal{B}^4(\Phi_T^1)+\mathcal{B}^3(\Phi_T^2),\\
\end{aligned}
\end{equation}
where $\mathcal{S}$ is sigmoid function. $\mathcal{B(\cdot)}$ denotes the bottleneck structure. $\mathcal{E(\cdot)}$ represents the $1 \times k$ deformable convolution in the horizontal direction and $k \times 1$ deformable convolution in the vertical direction.

The attentive features are then summed together:
\begin{equation}
\label{oaf_2}
	\Phi_T = \mathcal{T}(\Phi_H^1 + \Phi_V^1 + \Phi_H^2 + \Phi_V^2),
\end{equation} 
where $\mathcal{T}(\cdot)$ representing a base convolutional block that balances the fusion of different features. This two-orientation attention mechanism in OAF is pivotal in extracting cross-level features and refining modality-aware features. OAF is utilized to perform cross-level feature fusion. Simultaneously, the different multimodal classification results $Y_1$ and $Y_2$ are obtain by classifiers $\mathcal{C}$:
\begin{equation}
\label{eqc1c2}
	C_1 = \mathcal{C}(\Phi_T^1),C_2 = \mathcal{C}(\Phi_T^1).
\end{equation}
\subsubsection{Transformer Reconstruction Decoder}
In order to extract long-distance dependency features, we use a transformer encoder $\mathcal{V(\cdot)}$ based on ViT \cite{dosoViTskiy2020image} to explore global spectral dependencies from CNN-based encoding features $\Phi_T$ containing rich spatial information. This is formulated as: 
\begin{equation}
\label{ViTencoder}
	\Phi_V=\mathcal{V}(\Phi_T).
\end{equation}
To obtain the joint distinguish capacity from CNN and transformer structures, we complete the final classification by:
\begin{equation}
\label{eqc}
	C = \mathcal{C}(\Phi_T,\Phi_V).
\end{equation}
The decoder $\mathcal{D(\cdot)}$ reconstructs input modalities:
\begin{equation}
\label{re}
	R_e^{k},R_g^{k}=\mathcal{D}(\Phi_V),
\end{equation}
where the $\mathcal{D}$ enhances the fused representation using ViT as a structure.

\begin{figure}[htpb]
	\centering
	\includegraphics[width=0.9\linewidth]{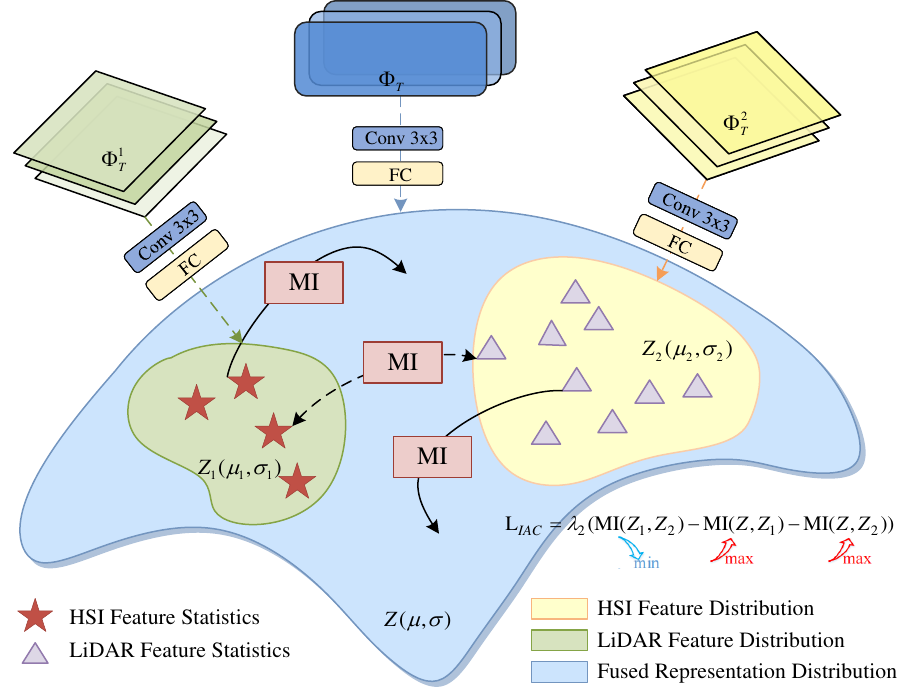}
	\caption{The illustration of complementary information preservation and redundancy information elimination.}
	\label{fig:MI}
 \vspace{-0.1in}
\end{figure}

\subsection{Information Aggregate-distributing Mechanism}

\textbf{\begin{figure*}[tp]
    \centering
    \includegraphics[width=\linewidth]{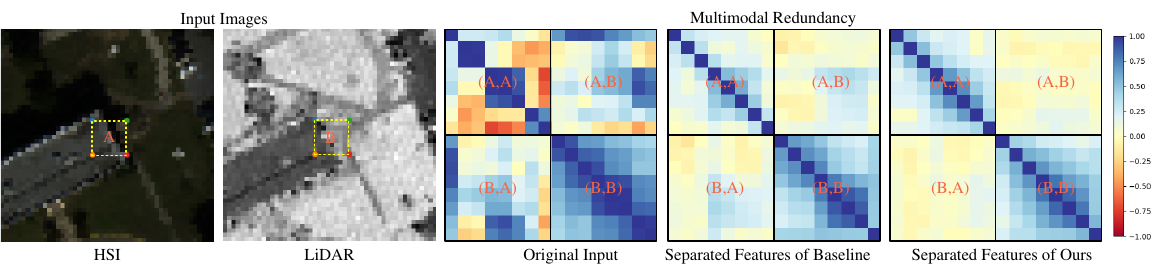}
    \caption{Visualization of the multimodal redundancy in the existing MLCC method and ours MIViT.}
    \label{fig:coco}
    \vspace{-0.1in}
\end{figure*}}

\subsubsection{Information Aggregation Constraint}
To tackle challenges of information redundancy and insufficient information in fusion maps, we employ mutual information maximization and minimization. We aim to enhance learning and fusion of complementary information between two modalities and reduce redundancy. The process of preserving complementary information and eliminating redundant information is illustrated in Fig. \ref{fig:MI}. Given unimodal features $\Phi_T^1$ and $\Phi_T^2$ and multi-modal fusion representation $\Phi_T$, we map them into low-dimensional vectors for mutual information analysis. An additional convolution layer with a kernel size of $3\times3$ and fully-connected layer is applied to map the above-reshaped features to obtain the probability distributions $Z_{1}$, $Z_{2}$ and $Z$:
\begin{equation}
	Z_{1} =\mathcal{M}(\Phi_T^1), Z_{2}=\mathcal{M}(\Phi_T^2),
	Z=\mathcal{M}(\Phi_T),
\end{equation}
where $\mathcal{M}_1$, $\mathcal{M}_2$ and $\mathcal{M}$ represent the convolution layers of kernel size $3 \times 3$ and fully-connected layers. In information theory, mutual information aims to measure the amount of information about a random variable by observing some other random variable or vice versa. The mutual information can be used to measure the difference between the entropy terms:
\begin{equation}
	\label{MI}
	\operatorname{MI}\left(Z, Z_{1}\right)=\operatorname{H}\left(Z \right)-\operatorname{H}\left(Z \mid Z_{1}\right),
\end{equation}
\begin{equation}
	\operatorname{MI}\left(Z, Z_{2}\right)=\operatorname{H}\left(Z \right)-\operatorname{H}\left(Z \mid Z_{2}\right).
\end{equation}
Take the calculation of Eq. \ref{MI} as an example:
\begin{equation}
	\operatorname{H}\left(Z \mid Z_{1}\right)=\operatorname{H}\left(Z,Z_{1}\right)-\operatorname{H}\left(Z_{1}\right),
\end{equation}
where $\operatorname{H}(\cdot)$ is the entropy, $\operatorname{H}(Z)$ and $\operatorname{H}(Z_{1})$ represents marginal entropies, and $\operatorname{H}(Z,Z_{1})$ indicate joint entropy of $Z$ and $Z_{1}$, $\operatorname{H}(Z,Z_{1})$ is the conditional entropy.
Combing above two equations,the mutual information $\operatorname{MI}\left(Z, Z_{1}\right)$ can be rewritten between the entropy terms:
\begin{equation}
	\operatorname{MI}\left(Z, Z_{1}\right)=\operatorname{H}\left(Z \right)+ \operatorname{H}\left(Z_{1} \right)-\operatorname{H}\left(Z, Z_{1}\right),
\end{equation}
where the marginal entropies $\operatorname{H}\left(Z \right)$ and $\operatorname{H}\left(Z_{1} \right)$ can be calculated with Kullback-Leibler divergence (KL) as:
\begin{equation}
	\operatorname{H}\left(Z \right) = \operatorname{H}_{Z_{1}} \left(Z \right) -KL(Z||Z_{1}),
\end{equation}
\begin{equation}
	\operatorname{H}\left(Z_{1} \right) = \operatorname{H}_{z} \left(Z_{1} \right) -KL(Z_{1}||Z),
\end{equation}
where $\operatorname{H}_{Z_{1}} \left(Z \right)$ is the cross-entropy. We then obtain:

\begin{equation}
\label{mi1}
	\begin{aligned}
		\operatorname{MI}\left(Z, Z_{1}\right)= & \operatorname{H}_{Z_{1}} \left(Z \right) + \operatorname{H}_{Z} \left(Z_{1} \right) - \operatorname{H}\left(Z, Z_{1}\right) \\ &-\operatorname{KL}(Z||Z_{1}) -\operatorname{KL}(Z_{1}||Z).
	\end{aligned}
\end{equation}
Similarly, we can calculate the mutual information between $Z$ and $Z_2$, as well as between $Z_1$ and $Z_2$:
\begin{equation}
\label{mi2}
	\begin{aligned}
		\operatorname{MI}\left(Z, Z_{2}\right)= & \operatorname{H}_{Z_{2}} \left(Z \right) + \operatorname{H}_{Z} \left(Z_{2} \right) - \operatorname{H}\left(Z, Z_{2}\right) \\ &-\operatorname{KL}(Z||Z_{2}) -\operatorname{KL}(Z_{2}||Z),
	\end{aligned}
\end{equation}
\begin{equation}
\label{mi3}
	\begin{aligned}
		\operatorname{MI}\left(Z_{1}, Z_{2}\right)= & \operatorname{H}_{Z_{2}} \left(Z_{1} \right) + \operatorname{H}_{Z_{1}} \left(Z_{2}\right) - \operatorname{H}\left(Z_{1}, Z_{2}\right) \\ &-\operatorname{KL}(Z_{1}||Z_{2}) -\operatorname{KL}(Z_{2}||Z_{1}).
	\end{aligned}
\end{equation}
Intuitively, $MI\left(Z_{1} Z_{2}\right)$ measures the reduction of uncertainty in $Z_{1}$ when $Z_{2}$ is observed, or vice versa while $\operatorname{MI}\left(Z, Z_{1}\right)$ and $\operatorname{MI}\left(Z, Z_{2}\right)$ measure the information transition from $Z_{1}$ and $Z_{2}$ to fusion features $Z$.

We use the correlation coefficient matrix to measure the redundancy between two modalities, as shown in Fig. \ref{fig:coco}. The original input images exhibit high correlation, but the redundancy of the features obtained through the neural network is reduced. This indicates that removing redundancy in multimodal learning is important for improving the network's ability to extract semantic information and recognize categories. Our algorithm has been validated to perform better in removing redundant information.

\subsubsection{Information Distribution Flow (IDF)}
IDF is designed to transfer task-relevant knowledge from deep to shallow classifiers, integrating performance-aware elements into fused representations and enhancing the classification ability of isolated modality classifiers. This design provides a solution to address missing multimodal learning scenarios. In a classification problem, a widely-used loss function between two discrete distributions is cross-entropy loss, whose gradient is equivalent to the gradient of KL divergence between two distributions. We assume $C$ is the deepest classifier and $C_n$ is the shallow classier output. Cross entropy loss from label to not only the deepest classifier but also all the shallow classier. It is computed with the labels from the training dataset and the outputs of each classifier's softmax layer. In this way, the knowledge hidden in the dataset is introduced directly from labels to all the classifiers:
\begin{equation}
\label{dis1}
	\begin{aligned}
\operatorname{D}(C_1,Y)+\operatorname{D}(C_2,Y) &= \operatorname{H}_{C_1}(Y) + \operatorname{H}_{C_2}(Y),\\
\operatorname{D}(C,Y) &= \operatorname{H}_{C}(Y).
	\end{aligned}
	\end{equation}
KL (Kullback-Leibler) divergence loss under the deepest classifier’s guidance. The KL divergence is computed using softmax outputs between the shallow classifiers and the deepest classifier and introduced to the softmax layer of each shallow classifier. By introducing KL divergence, the self-distillation framework affects the deepest classifier, the deepest one, to each shallow classifier:
\begin{equation}
\label{dis2}
			\operatorname{D}(C_1,C)+\operatorname{D}(C_2,C) =  \operatorname{KL}(C_1||C)+\operatorname{KL}(C_2||C).
	\end{equation}

\begin{figure}[htpb]
	\centering
	\includegraphics[width=0.9\linewidth]{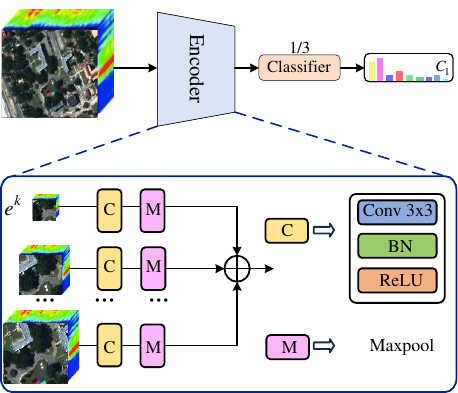}
	\caption{The illustration of single-modal shallow classifier for the missing multimodal learning in the inference stage.}
	\label{fig:missing}
 \vspace{-0.1in}
\end{figure}

The popular solutions to the problem of missing multimodal learning \cite{woo2023towards,ma2021smil,lee2023multimodal} are to design a cross-modality reconstruction network or separate the tasks of two modalities. Generally, it is difficult to reconstruct heterogeneous modalities from each other, while modal separation may lead to impaired information interaction. In this paper, the IDF is considered, which distributes the refined classification results of globally fused information from deep classifiers to shallow classifiers in different modal branches, enhancing their performance perception ability. This enables independent modal classification tasks to be achieved even in the absence of a certain modality, using a single-modal shallow classifier trained on a single-modal dataset as shown in Fig. \ref{fig:missing}. This is a lightweight CNN network, comprising only k (k set as 3 in paper) convolution layers, k maxpool layers, and a fully connected layer.

\begin{algorithm}
	\renewcommand{\algorithmicrequire}{\textbf{Input:}}
	\renewcommand{\algorithmicensure}{\textbf{Output:}}
	\caption{Training MIViT model for fusion classification of multimodal remote sensing data}
	\label{alg:1}
	\begin{algorithmic}[1]
		\REQUIRE Multimodal images $X_n$ (n=1,2), training label $Y$, training epochs \textit{epochs}, types of cubes $N$;
		\ENSURE Classification result $C$.
		\STATE Divide $X_n$ into contextual cubes $\{e_k\}_{k=1}^N$ and $\{g_k\}_{k=1}^N$.
        \STATE Initialize all weights.
		\WHILE {epoch $<$ \textit{epochs} do}
		\STATE Extract the separated features $\Phi_T^1$ and $\Phi_T^2$ from the alignment encoder by Eqs. \ref{ae1}, \ref{ae2} and \ref{ae3} .
		\STATE  Obtain the fused representation  $\Phi$ from oriented attention fusion by Eq. \ref{oaf_1} and Eq. \ref{oaf_2}.
		\STATE Generate the long-range information $\Phi_V$ from the ViT encoder by Eq. \ref{ViTencoder}.
        \STATE Acquire the classification maps $C_n$ and $C$ in the shallow and latent classifiers according to  Eq. \ref{eqc} and Eq. \ref{eqc1c2}, respectively.
        \STATE Measure the information redundancy among distributions of the separated features and fusion representations using mutual information entropy by Eqs. \ref{mi1}, \ref{mi2} and \ref{mi3}.
        \STATE Distillate the performance-aware classification probability of fused representation to the separated features according to Eq. \ref{dis1} and Eq. \ref{dis2}.
        \STATE Reconstruct cubes utilizing the ViT decoder in the way of Eq. \ref{re}.
        \STATE Update the MIViT model according to Eq. \ref{loss}.
		\ENDWHILE
	\end{algorithmic}  
\end{algorithm}

\subsection{Optimization}
\label{Optimization}
In this part, a variational optimization method is proposed to optimize the objective function of the MIViT method.
As shown in Fig. \textcolor{red}{\ref{fig:framenew}}, our final objective function is:
\begin{equation}
	\label{loss}
	\mathcal{L}= \mathcal{L}_{IDF}+\mathcal{L}_{Re}+ \mathcal{L}_{IAC}.
\end{equation}
And the Algorithm \ref{alg:1} clarifies the process of our MIViT method. The reconstruction losses ensure that the information contained in the images is not lost during the encoding and decoding process, \textit{i.e.}:
\begin{equation}
	\mathcal{L}_{\mathrm{Re}}=\lambda_1 \frac{1}{N} \sum_{k=1}^{N} \left(e^{k}-R_e^{k}\right)^2+\left(g^{k}-R_g^{k}\right)^2,
\end{equation}
where $\lambda_1$ is the tuning parameter. By maximizing $\operatorname{MI}\left(Z, Z_{1}\right)$ and $\operatorname{MI}\left(Z, Z_{2}\right)$ can explore the complementary attributes of both modalities and enhance the feature representation of the multi-modal fusion.
By minimizing $\operatorname{I}\left(Z_{1}, Z_{2}\right)$, we can effectively reduce the information redundancy between the two modalities. 
Hence the final IAC objective loss function is defined as:
\begin{equation}
	\mathcal{L}_{IAC}=\lambda_2
	(\operatorname{MI}(Z_1,Z_2)-\operatorname{MI}(Z,Z_1)-\operatorname{MI}(Z,Z_2)),
\end{equation}
where $\lambda_2$ is the tuning parameter. The IDF objective loss function is defined as:
\begin{equation}
	\begin{aligned}
\mathcal{L}_{IDF} = \lambda_3(&\operatorname{D}(C_1,Y)+\operatorname{D}(C_2,Y)) + \operatorname{D}(C,Y) \\ & +\lambda_4(\operatorname{D}(C_1,C)+\operatorname{D}(C_2,C)).
\end{aligned}
\end{equation}
where $\lambda_3$ as well as $\lambda_4$ are the tuning parameters.

\section{EXPERIMENTS}
\label{sec:exp}
\subsection{Datasets and metrics}
\label{}

In our study, we conducted experiments on three widely recognized benchmarks to assess the performance of our proposed fusion method: \textbf{Houston2013},  \textbf{MUUFL}, and \textbf{Trento}. Detailed characteristics of these datasets are presented in Table \ref{datadescripition}. To evaluate the classification outcomes quantitatively, we employed three key metrics: overall accuracy (OA), average accuracy (AA), and Kappa coefficient.

\begin{table}[htpb]
	\setlength{\tabcolsep}{1mm}
	\renewcommand{\arraystretch}{1.3}
	\centering
    \caption{Dataset Description.}
	\begin{tabular}{c|ccc}
        \hline
		Dataset      & Houston      & MUUFL     & Trento  \\ \hline
		Location    & Houston,Texas,USA    &\multicolumn{1}{c}{\begin{tabular}[c]{@{}c@{}}Long Beach,\\ Mississippi,USA\end{tabular}}   & Trento,Italy  \\
		Sensor Type     & HSI,DSM   & HSI,DSM,DEM & HSI,DSM \\
		Channel Number & 144,1   & 64,2  & 63,1    \\       
		Image Size (H)    & 349   & 325   &  600   \\
		Image Size (W)     & 1905 &  220  & 166  \\
		Train Samples & 2832  & 2683 & 819  \\
		Test Samples & 12197  &51004 & 29595  \\
		Classes & 15  &11 & 6   \\
		\hline
	\end{tabular}
    \label{datadescripition}
	\vspace{-0.1in}
\end{table}

\subsection{Implement details}
The experimental setup was executed on a system equipped with an NVIDIA GeForce RTX 3090 GPU. During the preprocessing phase, training samples were randomly cropped into patches of sizes $8 \times 8$, $16 \times 16$, and $24 \times 24$. For optimization, we utilized the Adam optimizer with an initial learning rate of 1e-4 and a weight decay set at 1e-3. We also employed a step scheduler with a step size of 50 and a gamma value of 0.9. The training process spanned 500 epochs. To align with most comparative experiments and ensure optimal performance, we set the batch size to 64. Regarding the network's hyperparameters, the decoder's configuration was mirrored to match that of the encoder. 

\begin{table*}[htpb]
	\centering
	\caption{OA, AA, and Kappa on the Houston2013 dataset by considering HSI and LiDAR data. The best result is \textbf{highlighted}.}
	\renewcommand{\arraystretch}{1.3}
	\setlength{\tabcolsep}{2mm}{
\begin{tabular}{ccccccccccccc}
\hline
 No. & Class(Train/Test)        & Endnet          & CoupleCNN       & Cross           & CALC            & CCR    & HCT             & ViT             & MFT            & ExViT & GLT             & MIViT           \\ \hline
1   & Healthy grass(198/1053)  & 83.86           & 83.10           & 83.10           & 78.63           & 88.6   & 82.91           & 82.59           & 82.34          & 81.2  & 83.10           & \textbf{91.07}  \\
2   & Stressed grass(190/1064) & 98.40           & 85.15           & 95.49           & 83.83           & 99.34  & 91.35           & 82.33           & 88.78          & 85.15 & 85.06           & 85.15           \\
3   & Synthetic grass(192/505) & \textbf{100.00} & 92.28           & 96.63           & 93.86           & 100.00 & \textbf{100.00} & 97.43           & 98.15          & 99.8  & 99.21           & 99.00           \\
4   & Tree(188/1056)           & 98.39           & 99.24           & 98.77           & 86.55           & 98.96  & 91.10           & 92.93           & 94.35          & 91.38 & 91.19           & \textbf{100.00} \\
5   & Soil(186/1056)           & 99.15           & \textbf{100.00} & \textbf{100.00} & 99.72           & 99.91  & \textbf{100.00} & 99.84           & 99.12          & 99.62 & \textbf{100.00} & \textbf{100.00} \\
6   & Water(182/143)           & 93.71           & 91.61           & 97.20           & 97.90           & 95.10  & 95.80           & 84.15           & 99.30          & 93.01 & 81.12           & 95.80           \\
7   & Residential(196/1072)    & 83.86           & 85.63           & 93.84           & 91.42           & 92.35  & 81.06           & 87.84           & 88.56          & 91.51 & 90.21           & 91.14           \\
8   & Commercial(191/1053)     & 91.17           & 72.74           & 91.93           & 92.88           & 90.98  & 94.97           & 79.93           & 86.89          & 97.44 & 96.87           & \textbf{97.44}  \\
9   & Road(193/1059)           & 85.08           & 90.75           & \textbf{95.28}  & 87.54           & 87.54  & 88.29           & 82.94           & 87.91          & 88.48 & 94.15           & 91.60           \\
10  & Highway(191/1036)        & 85.62           & 62.55           & 72.97           & 68.53           & 78.86  & 76.45           & 52.93           & 64.70          & 81.56 & 75.58           & 78.47           \\
11  & Railway(181/1054)        & 82.35           & 97.82           & 91.65           & 93.36           & 91.18  & 97.25           & 80.99           & \textbf{98.64} & 94.31 & 94.97           & 97.53           \\
12  & Park lot 1(192/1041)     & 83.57           & 94.43           & 87.61           & 95.10           & 82.04  & 91.55           & 91.07           & 94.24          & 93.76 & 97.89           & \textbf{98.94}  \\
13  & Park lot 2(184/285)      & 76.14           & 91.23           & 92.98           & 92.98           & 81.05  & 88.42           & 87.84           & 90.29          & 90.53 & 91.58           & 90.18           \\
14  & Tennis court(181/247)    & 100.00          & 87.85           & 93.12           & \textbf{100.00} & 93.93  & \textbf{100.00} & \textbf{100.00} & 99.73          & 97.57 & 99.19           & \textbf{100.00} \\
15  & Running track(187/473)   & 98.31           & 96.41           & \textbf{100.00} & 99.37           & 98.73  & 95.56           & 99.65           & 99.58          & 97.04 & 98.94           & 98.94           \\ \hline
    & OA(\%)                   & 89.93           & 87.91           & 91.84           & 88.97           & 91.56  & 90.43           & 85.05           & 89.80          & 91.27 & 91.64           & \textbf{93.71}  \\
    & AA(\%)                   & 90.64           & 88.72           & 92.70           & 90.78           & 91.90  & 91.65           & 86.83           & 91.54          & 92.16 & 91.94           & \textbf{94.35}  \\
    & Kappa(\%)                & 89.07           & 86.88           & 91.16           & 88.06           & 90.84  & 89.62           & 83.84           & 88.93          & 90.53 & 90.94           & \textbf{93.17}   \\ \hline
\label{com_result}
\vspace{-0.1in}
\end{tabular}}
\end{table*}

\begin{figure*}[htpb]
	\centering
	\includegraphics[width=\linewidth]{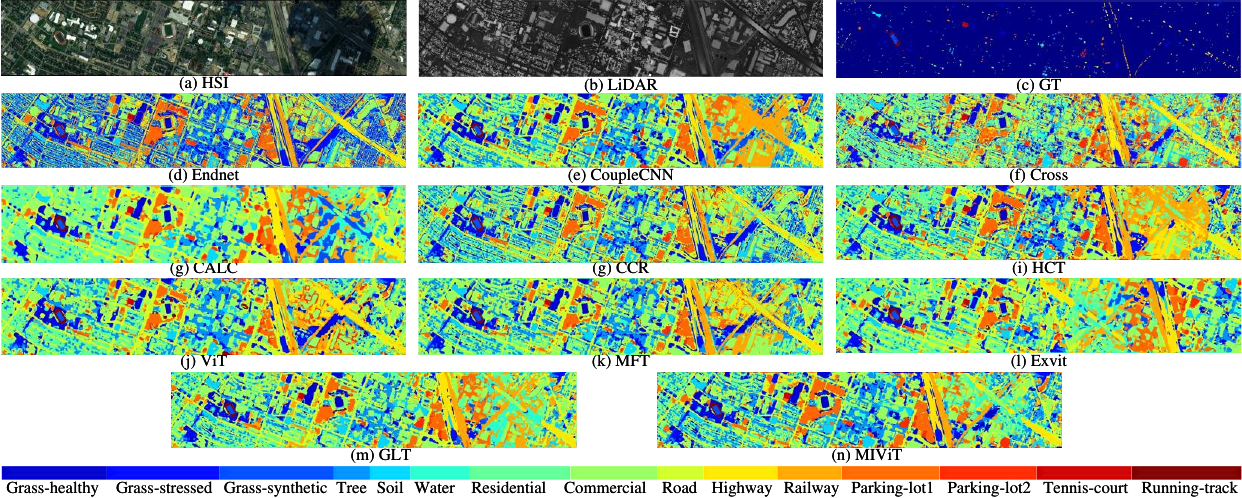}
	\caption{Classification maps for the Trento data obtained with different methods. (a) Pseudocolor image for HSI. (b) LiDAR-based DSM. (c) Ground-truth map. (d) Endnet (89.93\%). (e) CoupleCNN (87.91\%). (f) Cross (91.84\%). (g) CALC (88.97\%). (h) CCR (91.56\%). (i) HCT (90.43\%). (j) ViT (85.05\%). (k) MFT (89.80\%). (l) ExViT (91.27\%).GLT (91.64\%). (m) MIViT (93.71\%).}
	\label{fig:houston}
 \vspace{-0.1in}
\end{figure*}

\subsection{Comparison with state-of-the-art methods}
In assessing the performance of our proposed MIViT method, we selected several leading deep learning-based MLCC methods for comparison. These include \textbf{Endnet} \cite{hong2020deep}, \textbf{CoupleCNN} \cite{hang2020classification}, \textbf{Cross} \cite{hong2020more}, \textbf{CALC} \cite{lu2023coupled}, \textbf{CCR} \cite{wu2021convolutional}, \textbf{HCT} \cite{zhao2022joint}, \textbf{ViT} \cite{dosoViTskiy2020image}, \textbf{MFT} \cite{roy2023multimodal}, \textbf{ExViT}  \cite{yao2023extended}, \textbf{GLT} \cite{ding2022global}. 
\subsubsection{Quantitative Comparison}
The comparative results over five datasets are presented in Table \ref{com_result} and Table \ref{com_result1}, with the highest values emphasized in bold. Our method outperforms other advanced classification techniques across all remote sensing datasets. Specifically, compared to the second-best methods, our method shows average gains of 1.87$\%$, 0.46$\%$, and 0.14$\%$ in OA on the Houston2013, MUUFL, and Trento datasets, respectively. In addition to OA, we observed consistent improvements in other metrics, highlighting the distinctiveness of the fused features. The MIViT method distinguishes itself as the only approach capable of handling various dataset types, showcasing remarkable generalization capabilities and potential for MLCC tasks. It demonstrates SOTA classification performance on all six benchmark multi-modal datasets, surpassing all existing deep fusion strategies. This validates the effectiveness of the mutual information theory in feature-level fusion for multi-modal land cover classification.

\begin{table*}[htpb]
	\centering
	\caption{OA, AA, and Kappa on the MUUFL dataset by considering HSI and LiDAR data. The best result is \textbf{highlighted}.}
	\renewcommand{\arraystretch}{1.3}
	\setlength{\tabcolsep}{2mm}{
\begin{tabular}{ccccccccccccc}
\hline
No. & Class(Train/Test)              & Endnet  & CoupleCNN         & Cross          & CALC  & CCR   & HCT   & ViT   & MFT   & ExViT & GLT            & MIViT           \\  \hline
1   & Trees(1162/22084)              & 95.27  & 97.56          & \textbf{98.55} & 97.31 & 97.44 & 97.40 & 97.85 & 97.90 & 98.47 & 98.12          & 98.35           \\
2   & Mostly grass(214/4056)         & 75.52  & 86.19          & 65.85          & 93.00 & 77.93 & 90.31 & 76.06 & 92.11 & 90.34 & 90.41          & \textbf{92.46}  \\
3   & Mixed ground surface(344/6538) & 78.53  & 90.47          & 79.24          & 91.57 & 82.90 & 85.62 & 87.58 & 91.80 & 90.81 & \textbf{91.94} & 91.92           \\
4   & Dirt and sand(91/1735)         & 87.32  & \textbf{96.25} & 69.22          & 95.10 & 88.13 & 91.59 & 92.05 & 91.59 & 92.39 & 94.52          & 94.81           \\
5   & Road(334/6353)                 & 94.16  & \textbf{97.37} & 97.12          & 95.91 & 96.49 & 94.66 & 94.71 & 95.60 & 93.69 & 95.73          & 94.13           \\
6   & Water(23/443)                  & 90.07  & 95.03          & 60.95          & 99.32 & 94.36 & 97.52 & 82.02 & 88.19 & 90.97 & 95.49          & \textbf{100.00} \\
7   & Building shadow(112/2121)      & 69.31  & 88.21          & 64.55          & 92.69 & 83.36 & 86.28 & 87.11 & 90.27 & 91.37 & 93.16          & \textbf{93.82}  \\
8   & Building(312/5928)             & 93.35  & 98.25          & 90.30          & 98.45 & 97.93 & 96.66 & 97.60 & 97.26 & 98.11 & 98.21          & \textbf{98.41}  \\
9   & Sidewalk(69/1316)              & 64.82  & 42.17          & 32.75          & 51.60 & 52.89 & 55.62 & 57.83 & 61.35 & 66.57 & 74.39          & \textbf{81.31}           \\
10  & Yellow curb(9/174)             & 51.72  & 3.45           & 0.00           & 0.00  & 2.30  & 21.26 & 31.99 & 17.43 & 32.76 & 28.74          & \textbf{56.32}           \\
11  & Cloth panels(13/256)           & 92.19  & 76.17          & 43.36          & 0.00  & 85.16 & 69.14 & 58.72 & 72.79 & 85.94 & 92.97          & \textbf{98.05}           \\  \hline
    & OA(\%)                         & 88.85  & 93.50          & 87.29          & 93.94 & 91.50 & 92.76 & 92.15 & 94.34 & 94.53 & 95.20          & \textbf{95.66}  \\
    & AA(\%)                         & 81.11  & 79.24          & 63.81          & 74.09 & 78.08 & 80.55 & 78.50 & 81.48 & 84.67 & 86.70          & \textbf{90.87}  \\
    & Kappa(\%)                      & 85.25  & 91.40          & 82.75          & 92.00 & 88.74 & 90.42 & 89.56 & 92.51 & 92.76 & 93.67          & \textbf{94.27}   \\ \hline
\label{com_result1}
\vspace{-0.1in}
\end{tabular}}
\end{table*}

\begin{figure*}[htpb]
	\centering
	\includegraphics[width=\linewidth]{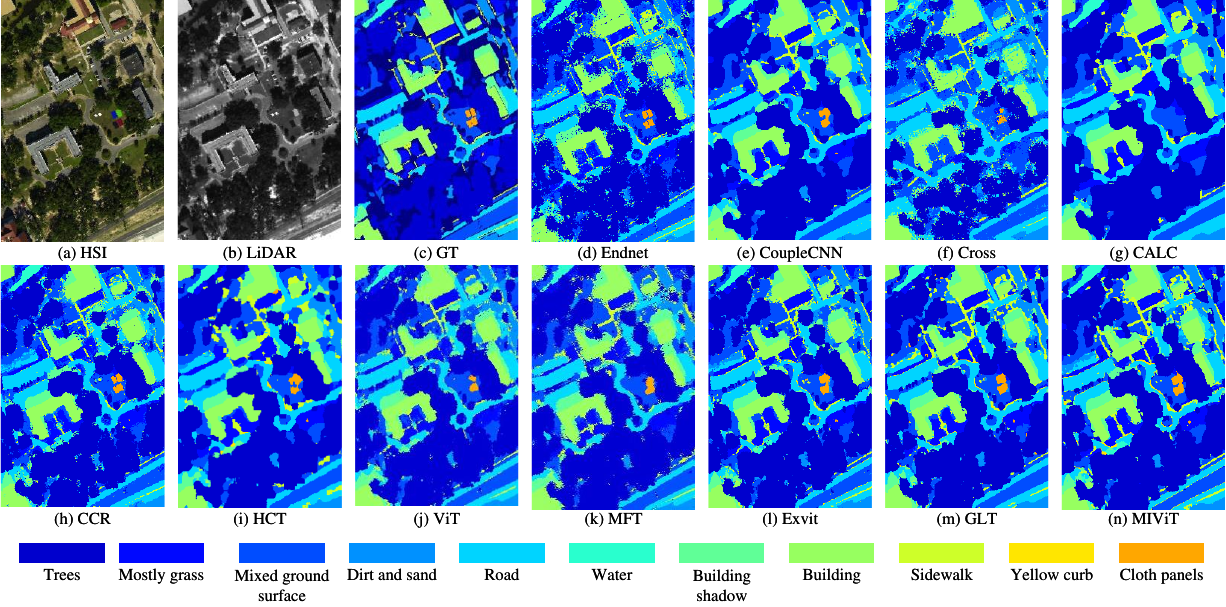}
 \vspace{-0.1in}
	\caption{Classification maps for the MUUFL data obtained with different methods. (a) Pseudocolor image for HSI. (b) LiDAR-based DSM. (c) Ground-truth map. (d) Endnet (88.85\%). (e) CoupleCNN (93.50\%). (f) Cross (87.29\%). (g) CALC (93.94\%). (h) CCR (91.50\%). (i) HCT (92.76\%). (j) ViT (92.15\%). (k) MFT (94.34\%). (l) ExViT (94.53\%).GLT (95.20\%). (m) MIViT (95.66\%).}
	\label{fig:MUUFL}
 \vspace{-0.1in}
\end{figure*}

\subsubsection{Qualitative Comparison}
\begin{figure*}[htpb]
	\centering
	\includegraphics[width=\linewidth]{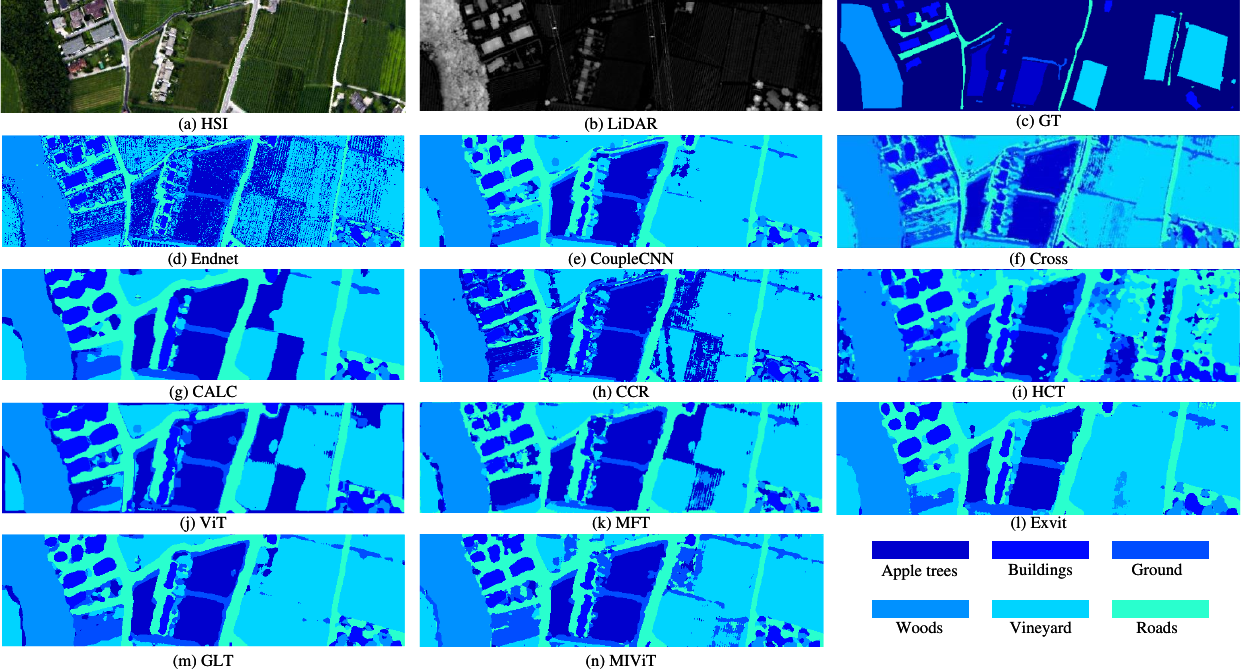}
	\caption{Classification maps for the Trento data obtained with different methods. (a) Pseudocolor image for HSI. (b) LiDAR-based DSM. (c) Ground-truth map. (d) Endnet (90.40\%). (e) CoupleCNN (98.82\%). (f) Cross (97.82\%). (g) CALC (98.11\%). (h) CCR (96.37\%). (i) HCT (86.72\%). (j) ViT (94.62\%). (k) MFT (97.76\%). (l) ExViT (98.58\%).GLT (99.29\%). (m) MIViT (99.43\%).}
	\label{fig:trento}
 \vspace{-0.1in}
\end{figure*}

We conducted a qualitative evaluation by visualizing the classification maps generated by different methods. Figs. \ref{fig:houston}, \ref{fig:MUUFL}, and \ref{fig:trento} display classification maps for hyperspectral datasets (Houston, MUUFL, Trento) alongside their corresponding LiDAR modalities. CNN-based models, due to their learnable feature extraction capabilities, capture nonlinear relationships between input and output feature maps, producing relatively smooth classification maps. Transformer-based models, like ViT, MFT, and ExViT, excel in hyperspectral classification by extracting high-level abstract representations, resulting in superior visual quality in classification maps compared to CNNs. The MIViT, by more efficiently reducing redundancy between modalities and enhancing the perception of independently extracted modality features, produces highly accurate classification maps, particularly when compared to ViT. This indicates that MIViT facilitates more effective information interaction, allowing for better fusion of spectral data from HSI with elevation data from LiDAR.

On the MUUFL dataset, as shown in Fig. \ref{fig:MUUFL}, our MIViT method achieves the best visual performance, closely resembling the ground truth map. It effectively utilizes high-level semantic information, particularly in the classification of Trees and Yellow curb, which appear more continuous compared to other methods. Notably, our method avoids issues like the omission of the Cloth panels category, a problem observed in the CALC method.

Fig. \ref{fig:trento} offers a similar perspective, displaying classification maps for different methods on the Trento dataset. The classification results from MIViT exhibit smoother class transitions and less noise. Misclassification, a common challenge in such tasks, is markedly reduced in our method. Categories like Apple trees and Vineyard, often prone to confusion, are more accurately classified by our method. While other methods show misclassifications in categories like Roads, MIViT achieves completely accurate results. Additionally, in the Ground category, our method's classification performance is significantly superior to other state-of-the-art classifiers.

\subsection{Ablation experiments}

In this section, we conducted ablation studies to further validate the effectiveness of our model's components. Specifically, we designed four scenarios to analyze the impact of key modules: (1) Exclusion of the Oriented Attention Fusion (OAF) and its replacement with a simple addition operation; (2) Removal of the Transformer Reconstruction Decoder (TRD) in the reconstruction stage; (3) Exclusion of the Information Aggregation Constraint (IAC) from the aggregation training of separated features and fused representations; (4) Omission of distillation from the information distribution flow (IDF) in the distribution training.

\begin{table*}[htpb]
	\centering
	\caption{OA, AA, and Kappa on the Trento dataset by considering HSI and LiDAR data. The best result is \textbf{highlighted}.}
	\renewcommand{\arraystretch}{1.3}
	\setlength{\tabcolsep}{2mm}{
\begin{tabular}{ccccccccccccc}
\hline
No. & Class(Train/Test)     & Endnet & CoupleCNN       & Cross          & CALC            & CCR    & HCT             & ViT   & MFT   & ExViT  & GLT             & MIViT           \\ \hline
1   & Apple trees(129/3905) & 92.88  & 98.98           & \textbf{99.87} & 98.62           & 100.00 & 99.26           & 89.07 & 98.20 & 99.13  & 99.23           & 99.59           \\
2   & Buildings(125/2778)   & 97.05  & 96.83           & 98.63          & \textbf{99.96}  & 98.67  & 99.96           & 84.74 & 91.65 & 98.56  & 98.74           & 99.53           \\
3   & Ground(105/374)       & 96.52  & 95.45           & 98.93          & 72.99           & 90.37  & \textbf{100.00} & 92.25 & 94.92 & 77.81  & 99.20           & \textbf{100.00} \\
4   & Woods(154/8969)       & 99.30  & 99.87           & 99.22          & \textbf{100.00} & 99.91  & \textbf{100.00} & 99.63 & 99.45 & 100.00 & \textbf{100.00} & \textbf{100.00} \\
5   & Vineyard(184/10317)   & 80.11  & \textbf{100.00} & 98.05          & 99.44           & 94.76  & 67.47           & 98.23 & 99.72 & 99.92  & \textbf{100.00} & 99.85           \\
6   & Roads(122/3052)       & 89.02  & 93.77           & 89.42          & 88.76           & 85.39  & 83.06           & 84.04 & 91.52 & 91.78  & 95.35           & \textbf{95.97}  \\ \hline
    & OA(\%)                & 90.40  & 98.82           & 97.82          & 98.11           & 96.37  & 86.72           & 94.62 & 97.76 & 98.58  & 99.29           & \textbf{99.43}  \\
    & AA(\%)                & 92.48  & 97.48           & 97.35          & 93.30           & 94.85  & 91.63           & 91.33 & 95.91 & 94.53  & 98.75           & \textbf{99.16}  \\
    & Kappa(\%)             & 87.35  & 98.42           & 97.09          & 97.46           & 95.16  & 82.95           & 92.81 & 97.00 & 98.10  & 99.04           & \textbf{99.24} \\ \hline
\label{com_result2}
 \vspace{-0.1in}
\end{tabular}}
\end{table*}

\begin{figure*}[htpb]
	\centering
	\includegraphics[width=0.85\linewidth]{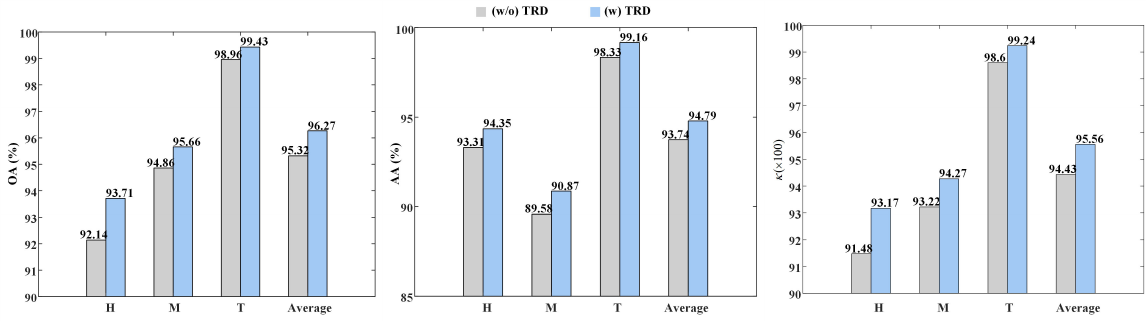}
	\caption{Effects of transformer reconstruction decoder (TRD) on the Houston2013(H), MUUFL(M), Trento(T) datasets.}
	\label{fig:TRD}
 \vspace{-0.1in}
\end{figure*}

\textbf{Effect of transformer reconstruction decoder module.} We integrated a feature reconstruction transformer encoder (TRD) into our fusion process to minimize information loss from different modalities during fusion, as illustrated in Fig. \ref{fig:framenew}. To assess TRD's effectiveness in feature recognition tasks, we conducted ablation experiments, using final classification accuracy as the metric. Fig. \ref{fig:TRD} evaluates the impact of TRD on OA, AA, and Kappa statistics across three datasets. The inclusion of TRD led to improvements in all performance metrics. This suggests that focusing more on discriminative spatial features is beneficial for improving classification accuracy, particularly for HSI and LiDAR data. Notably, the most significant improvement was observed for the Houston dataset, with AA improved by nearly 1.6\%. The Houston dataset contains a wide variety of complex ground object categories, resulting in significant variations in the probabilities for each classification. In such cases, TRD can have a more pronounced effect. This highlights TRD's significant role in enhancing classification accuracy, especially for complex ground object categories. 

\begin{figure*}[htpb]
	\centering
	\includegraphics[width=0.85\linewidth]{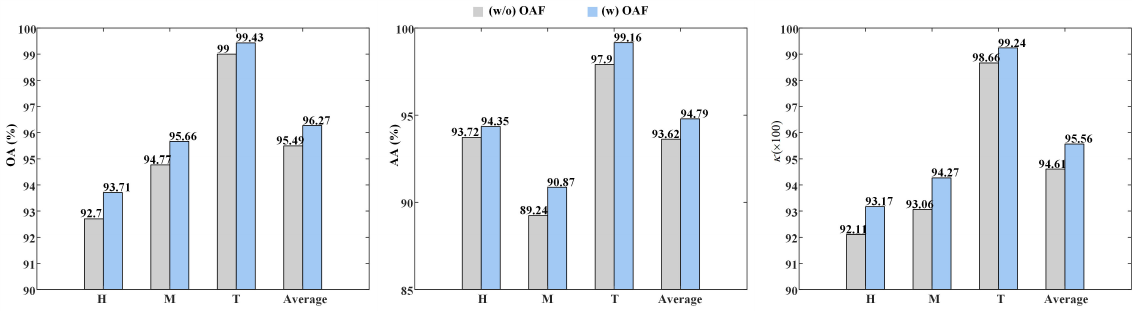}
	\caption{Effects of oriented attention fusion (OAF) on the Houston2013(H), MUUFL(M), Trento(T) datasets.}
	\label{fig:OAF}
 \vspace{-0.1in}
\end{figure*}

\textbf{Effect of oriented attention fusion module.} As shown in Fig. \ref{fig:framenew}, we incorporated oriented attention fusion (OAF) to facilitate the effective extraction and fusion of multi-layer features from different modalities, allowing for information interaction in various directions. To demonstrate OAF's effectiveness, we conducted ablation experiments, replacing OAF with a simple addition operation. Fig. \ref{fig:OAF} shows that the inclusion of OAF led to improvements in overall performance metrics across all datasets, confirming its advantage in enhancing model performance.

\begin{figure}[htpb]
	\centering
	\includegraphics[width=\linewidth]{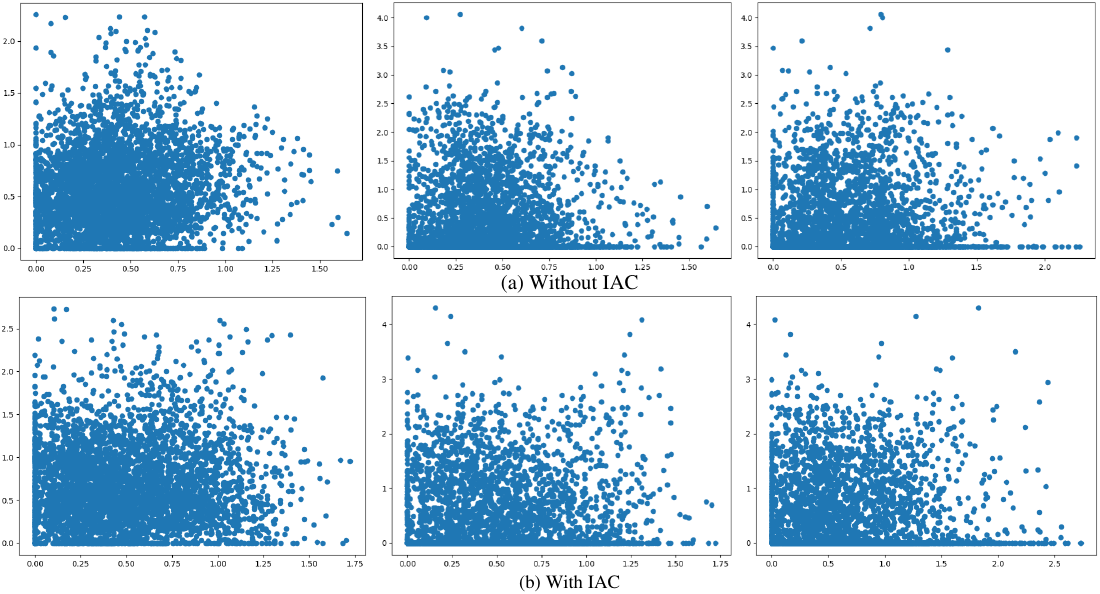}
	\caption{The distribution visualization of correlation relationship between the separated features from different modalities.}
	\label{fig:sca}
 \vspace{-0.1in}
\end{figure}

\begin{figure}[htpb]
	\centering
	\includegraphics[width=\linewidth]{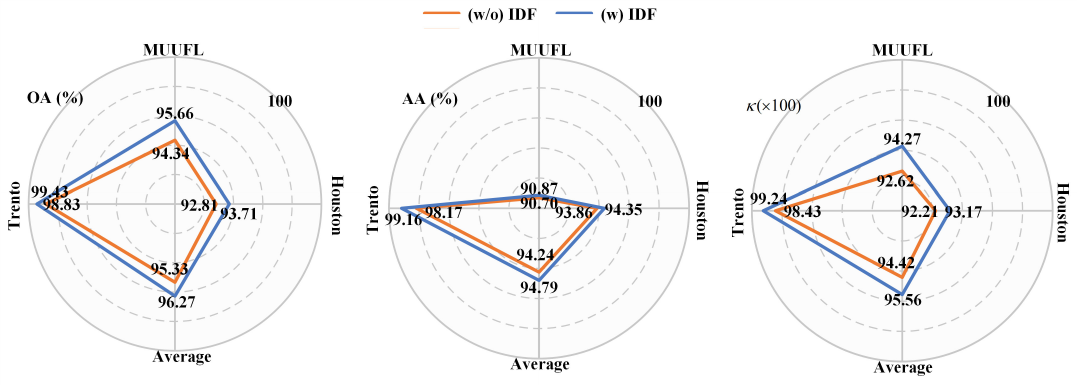}
	\caption{Effects of information aggregation constraint (IAC) on the Houston2013, MUUFL, Trento datasets.}
	\label{fig:IAC}
 \vspace{-0.1in}
\end{figure}

\textbf{Effect of information aggregation constraint module.} Fig. \ref{fig:framenew} depicts our introduction of IAC to reduce redundancy between modalities and ensure consistent information in fusion representations. Ablation experiments assessing the impact of IAC on OA, AA, and Kappa are presented in Fig. \ref{fig:IAC}. The addition of IAC led to enhancements in all metrics. This indicates that paying more attention to discriminative spatial features contributes to improved classification accuracy. The improvement on AA is relatively pronounced, especially for the Trento dataset, as IAC effectively alleviates the phenomenon of non-uniform classification difficulty. In such cases, IAC can play a more significant role. These results suggest that IAC indeed makes a positive contribution to the classification task. To visualize the efficacy of the IAC module in eliminating redundancy, we generated scatter plots for features extracted from different modalities. A more scattered distribution indicates a lower correlation between two features, implying less intermingled redundant information. Upon inspecting Fig \ref{fig:sca}, it is evident that the model incorporating the IAC module exhibits a more dispersed distribution among different features. This observation suggests that the IAC module plays a crucial role in effectively reducing redundancy in multimodal information. 

\begin{figure}[htpb]
	\centering
	\includegraphics[width=\linewidth]{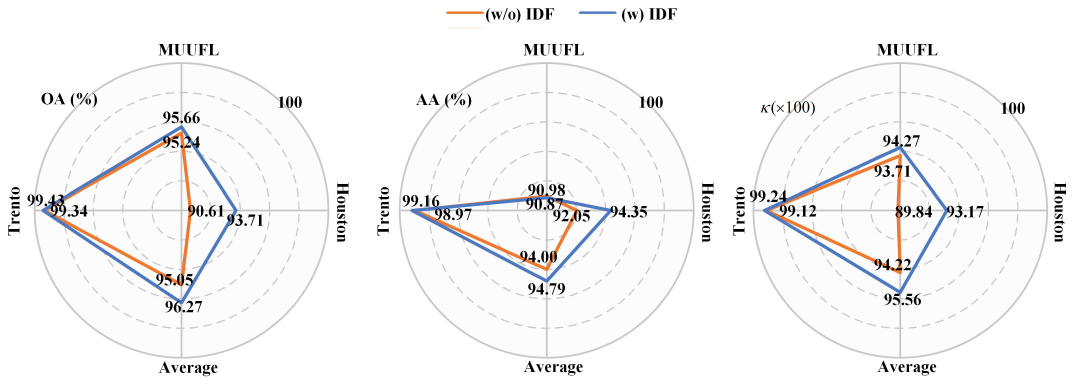}
	\caption{Effects of information distribution flow (IDF) on the Houston2013, MUUFL, Trento datasets.}
	\label{fig:IDF}
 \vspace{-0.1in}
\end{figure}

\textbf{Effect of information distribution flow module.}
As illustrated in Eq. \ref{eqc}, the final classification result is achieved through the decision fusion of CNN and ViT encoders, aiming for a globally optimal classification outcome. Individual modality-specific classification results are computed using Eq. \ref{eqc1c2}. The globally optimal classification result exhibits strong performance awareness and, through the IDF loss based on self-distillation, can enhance the performance awareness of individual modalities by distributing information across different modalities. The IDF module's effectiveness was validated through ablation experiments, with the impact evaluated in terms of OA, AA, and Kappa (Fig. \ref{fig:IDF}). With the inclusion of the IDF module, shallow, individual modality features are paired with their respective classifiers. When the IDF module is omitted, these classifiers are removed. The inclusion of IDF improved all metrics, demonstrating its role in enhancing perceptual capabilities of separated modality features and thereby strengthening fusion representation classification results. This confirms IDF's positive impact on the classification task.

%
\subsection{Missing Multimodal Learning}
\begin{figure}[htpb]
	\centering
	\includegraphics[width=\linewidth]{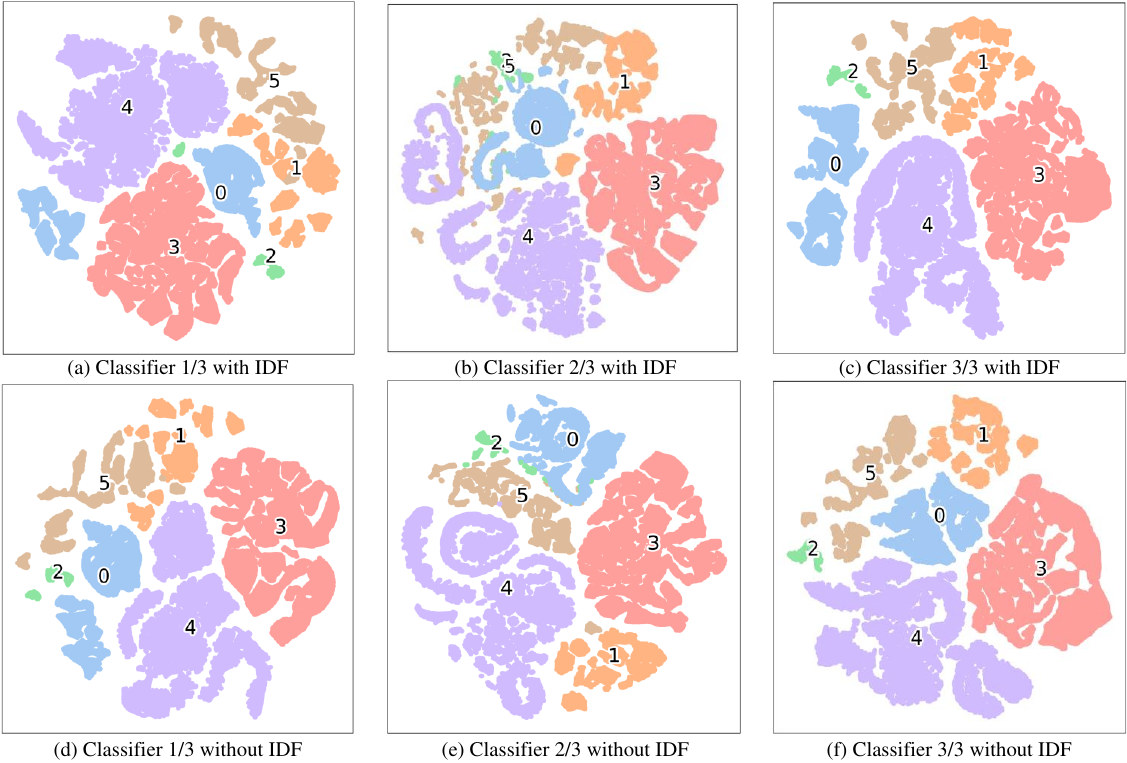}
	\caption{tSNE visualization of feature distribution in different classifiers.}
	\label{fig:tsne}
 \vspace{-0.1in}
\end{figure}

Fig. \ref{fig:tsne} presents an analysis of the features across different classifiers. Classifiers 1/3 and 2/3 represent features of shallow separated modalities, while Classifier 3/3 denotes the latent fused representation. A notable observation is the enhanced discriminative performance of classifiers incorporating the Information Distribution Flow (IDF) compared to those without it. Specifically, Class 2 shows a clear distinction from other classes in the two-dimensional projection space. This underscores the significant accuracy enhancement achieved through IDF, emphasizing its crucial role in land cover classification tasks. 

\begin{table}[htpb]
	\centering
	\caption{Accuracy measurement of the different classifiers (classifier 1/3, 2/3 and 3/3) in MIViT. The best result of each modality is \textbf{highlighted}.}
	\renewcommand{\arraystretch}{1.3}
	\setlength{\tabcolsep}{0.2mm}{
\begin{tabular}{cc|ccc|ccc|ccc}
\hline
                        &             & \multicolumn{3}{c|}{Houston} & \multicolumn{3}{c|}{MUUFL} & \multicolumn{3}{c}{Trento} \\
Modality             & Method                                                        & OA      & AA      & Kappa   & OA      & AA     & Kappa  & OA      & AA      & Kappa  \\ \hline
\multirow{2}{*}{HSI}               & Cross                                                         & 83.73   & 86.62   & 82.41   & 86.22   & 64.72  & 81.69  & 87.59   & 87.32   & 83.68  \\
                                   & 1/3 & \textbf{88.47}   & \textbf{89.88}   & \textbf{87.49}   & \textbf{92.43}   & \textbf{82.92}  & \textbf{89.99}  & \textbf{96.87 }  & \textbf{94.94}   & \textbf{95.80}  \\ \hline
\multirow{2}{*}{LiDAR}               & Cross                                                         & 62.01   & 63.45   & 58.90   & \textbf{70.19}   & 39.65  & \textbf{59.84}  & 84.74   & 82.95   & 80.24  \\
                             &  2/3 & \textbf{64.49 }  & \textbf{65.84}   & \textbf{61.47}   & 69.41   & \textbf{43.15}  & 57.56  & \textbf{95.60}   & \textbf{87.54}   & \textbf{94.15}  \\ \hline
\multirow{3}{*}{Fusion}           & Cross                                                         & 91.84   & 92.70   & 91.16   & 87.29   & 63.81  & 82.75  & 97.82   & 97.35   & 97.09  \\
                      & MFT                                                           & 89.80   & 91.51   & 88.93   & 94.34   & 81.48  & 92.51  & 97.76   & 95.91   & 97.00  \\
                        &  3/3 & \textbf{93.71}   & \textbf{94.35}   & \textbf{93.17 }  & \textbf{95.66}   & \textbf{90.87}  & \textbf{94.27}  & \textbf{99.43}   &\textbf{ 99.16}   &\textbf{ 99.24} \\ \hline
\label{incomplete}
\vspace{-0.1in}
\end{tabular}}
\end{table}
Table \ref{incomplete} summarizes the accuracy of each classifier, revealing that the global classifier (3/3) achieves higher classification accuracy — 96.15\% OA, 91.01\% AA, and 95.04\% Kappa — compared to the single-modality classifiers (1/3 and 2/3). This indicates that multimodal feature-level fusion markedly improves accuracy, highlighting its significance in remote sensing tasks. However, the classifier for HSI (1/3) records suboptimal accuracy, likely due to its wide spectral coverage and diverse spectral information.
When compared with Cross \cite{hong2020more}, which specifically tackles missing multimodal learning, our method shows substantial improvements across three datasets and various modalities. For instance, on the Trento dataset, our approach achieved a 9.28\% increase in OA, 7.35\% in AA, and 12.12\% in Kappa for the HSI modality. In the LiDAR modality, the improvements are even more pronounced, with increases of 10.86\% in OA, 4.59\% in AA, and 5.08\% in Kappa. This demonstrates that our algorithm effectively channels global fusion information into modalities, thereby enhancing the performance of single-modal detectors.
Moreover, we evaluated the complexity and computational load of different modal detectors within our algorithm, as shown in Table \ref{para}. Parameters (Paras) indicate the model size, while FLOPs measure computational complexity. The high-dimensional characteristics of HSI require more parameters and computational resources than LiDAR data. Our study introduces an innovative approach for addressing missing modalities. By using a global deep detector to distill information from shallow classifiers, we enhance the extraction of detailed information during the feature extraction process of shallow single modalities and augment semantic information. In cases of missing modalities, the classification task can still be effectively accomplished using the feature extractor and classifier from another modality.

\begin{table}[htpb]
	\centering
	\caption{Comparison of model complexity of different modalities on all three datasets.}
	\renewcommand{\arraystretch}{1.3}
	\setlength{\tabcolsep}{4mm}{
\begin{tabular}{cccc}
\hline
Modality & Method        & Params/K & FLOPs/M \\ \hline
HSI        & Classifier 1/3 & 84.48    & 80.45   \\
LiDAR      & Classifier 2/3 & 58.85    & 34.51   \\
Fusion     & Classifier 3/3 & 1867.00  & 900.12 \\ \hline
\label{para}
\vspace{-0.1in}
\end{tabular}}
\end{table}

\section{Conclusions}
\label{sec:conclusion}
In summary, our proposed Multimodal Informative ViT (MIViT) offers a highly effective solution for addressing empirical redundancy in multimodal land cover classification (MLCC). MIViT integrates an innovative information aggregate-distributing mechanism, redefining the redundancy level and enhancing performance-awareness in fused representations. This approach promotes a more efficient integration of complementary features from each modality. Central to MIViT's effectiveness is its bidirectional learning process, encompassing both forward and backward directions, which enables the fused representation to comprehensively capture semantic information. The incorporation of Oriented Attention Fusion (OAF) and a Transformer feature extractor is pivotal in extracting both shallow local and deep global features, respectively. Furthermore, the implementation of an Information Aggregation Constraint (IAC), grounded in mutual information, adeptly reduces redundant information while preserving the correlation of complementary information. The Information Distribution Flow (IDF) augments the performance-awareness of individual modalities, distributing global classification information across various modality feature maps. This sophisticated architecture is also adept at handling missing modality challenges, utilizing lightweight independent modality classifiers. Our extensive experimental analyses validate the superiority of MIViT, demonstrating its remarkable improvements over existing state-of-the-art methods across five multimodal datasets. Achieving an overall accuracy of 95.56\%, MIViT establishes itself as a robust and innovative approach for multimodal land cover classification, effectively bridging the gap between multiple data sources and accurate, efficient classification.

\newpage
	\bibliographystyle{IEEEtran}
	\bibliography{refs}

\end{document}